\documentclass[letterpaper, 10 pt, conference]{ieeeconf}  

\usepackage{blindtext, graphicx}
\usepackage{graphicx}
\usepackage{xcolor}
\usepackage{amsmath}
\usepackage{cite}
\usepackage[english]{babel}
\usepackage{amsfonts}
\usepackage{verbatim}
\usepackage{epsfig} 
\usepackage{psfrag}
\usepackage{amssymb}
 \usepackage{cases}
\usepackage{gensymb}
\usepackage{tikz}
\usepackage{textcomp}
\usepackage{hyperref}
\usepackage{lipsum}

\newtheorem{lemma}{Lemma}
\newtheorem{theorem}{Theorem}
\newtheorem{definition}{Def.}

\newtheorem{assumption}{Assumption}
\newtheorem{remark}{Remark}

\newcommand{\ad}{\text{ad}}
\newcommand{\Ad}{\text{Ad}}

\newcommand{\tr}{\text{tr}}

\newcommand{\FontFigXS}{.6}

\newcommand{\FontFigS}{.8}
\newcommand{\FontFigM}{1}

\newcommand*{\QEDB}{\hfill\ensuremath{\square}}
\newcommand\copyrighttext{%
  \footnotesize \textcopyright 2019 IEEE.  Personal use of this material is permitted.  Permission from IEEE must be obtained for all other uses, in any current or future media, including reprinting/republishing this material for advertising or promotional purposes, creating new collective works, for resale or redistribution to servers or lists, or reuse of any copyrighted component of this work in other works.
  }
\newcommand\copyrightnotice{%
\begin{tikzpicture}[remember picture,overlay]
\node[anchor=south,yshift=10pt] at (current page.south) {\fbox{\parbox{\dimexpr\textwidth-\fboxsep-\fboxrule\relax}{\copyrighttext}}};
\end{tikzpicture}%
}

\IEEEoverridecommandlockouts                              

\title{\LARGE \bf
A Nonlinear Observer for Free-Floating Target Motion\\ using only Pose Measurements
}

\author{Hrishik Mishra$^{1}$, Marco De Stefano$^{1}$, Alessandro Massimo Giordano$^{1,2}$ and Christian Ott$^{1}$
\thanks{$^{1}$The author is with Institute of Robotics and Mechatronics, German Aerospace Center (DLR), We{\ss}ling, Oberpfaffenhofen. $^{2}$The author is with Technical University of Munich, Dept. of Informatics, Germany.
        {Contact e-mail: \tt\small hrishik.mishra@dlr.de}}%
}

\begin{document}

\maketitle
\copyrightnotice
\thispagestyle{empty}
\pagestyle{empty}

\begin{abstract}
In this paper, we design a nonlinear observer to estimate the inertial pose and the velocity of a free-floating non-cooperative satellite (Target) using only relative pose measurements. In the context of control design for orbital robotic capture of such a non-cooperative Target, due to lack of navigational aids, only a relative pose estimate may be obtained from slow-sampled and noisy exteroceptive sensors. The velocity, however, cannot be measured directly. To address this problem, we develop a model-based observer which acts as an \emph{internal model} for Target kinematics/dynamics and therefore, may act as a predictor during periods of no measurement. To this end, firstly, we formalize the estimation problem on the $\text{SE}(3)$ Lie group with different state and measurement spaces. Secondly, we develop the kinematics and dynamics observer such that the overall observer error dynamics possesses a stability property. Finally, the proposed observer is validated through robust Monte-Carlo simulations and experiments on a robotic facility.
\end{abstract}

\section{Introduction}

\begin{figure}[h]
  \centering  	
    \psfrag{I}[cc][cc][\FontFigM]{{\color{black}$\{\mathcal{I}\}$}}
    \psfrag{E}[cc][cc][\FontFigM]{{\color{black}$\{\mathcal{E}\}$}}
    \psfrag{T}[cc][cc][\FontFigM]{{\color{black}$\{\mathcal{T}\}$}}
    \psfrag{G}[cc][cc][\FontFigM]{{\color{black}$\{\mathcal{G}\}$}}
    \psfrag{g}[cc][cc][\FontFigM]{{\color{black}$g$}}
    \psfrag{gtg}[cc][cc][\FontFigM]{{\color{black}$g_{tg}$}}
    \psfrag{ge}[cc][cc][\FontFigM]{{\color{black}$g_e$}}
    \psfrag{gt}[cc][cc][\FontFigM]{{\color{black}$g_t$}}
    \includegraphics[width=0.35\textwidth]{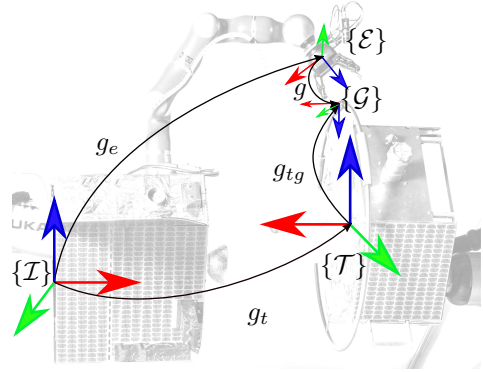}
    \caption{Kinematic description of an OOS-SIM scenario in which a manipulator-equipped spacecraft (left) tracks a free-floating satellite (Target, right). The grasping frame $\{\mathcal{G}\}$ is observed in an end-effector-mounted camera frame, $\{\mathcal{E}\}$, and the motion states $\{g_t,V_t^b\}$ with respect to inertial frame $\{\mathcal{I}\}$ have to be estimated.} \label{fig_rigid} 
\end{figure}

The estimation of motion parameters is key to several Cartesian control methods for robots and vehicles. For regulation problems, a pose estimate using a kinematic observer is sufficient. However, for tracking problems of the kind where the motion of the desired frame is time-varying, an additional velocity measurement is required in the commonly known PD+ controllers \cite{Paden}. In the context of On-Orbit Servicing (OOS, see Fig.~\ref{fig_rigid}), the control objective is to track a grasping frame on a free-floating satellite (right) with a manipulator-equipped spacecraft (left). In the specific case that such a satellite is also non-cooperative (Target), the available measurement is often a relative pose which is computed using an exteroceptive sensor like a camera. Therefore, the control design is negatively affected by the lack of in-situ proprioceptive sensors, like an Inertial Measurement Unit (IMU), to measure the Target's velocity, thereby motivating the need for an observer of the Target's motion states.

Pertaining only to attitude estimation, in \cite{markley}, the Multiplicative Extended Kalman Filter (M-EKF) was proposed which dealt with the orientation manifold by performing measurement update in the tangent space of the quaternion group. In \cite{Bonnabel2}, the authors developed a theoretical treatment of Lie group observers that adhered to symmetries in kinematics and possessed autonomous error dynamics. In \cite{Bonnable} and \cite{Barrau}, this foundation was used to develop an invariant EKF which was proved to be locally exponentially stable. In \cite{MAHONY2013617},  the invariant observer theory was enhanced and the autonomous error dynamics derived in \cite{Bonnabel2} was corroborated. The approaches by both \cite{Bonnabel2} and \cite{Mahony} propose an internal model (pre-observer) which possesses a geometric structure which is identical to the actual kinematic system. In \cite{Bourmaud2015}, the Continuous-Discrete-EKF was developed which formalized filtering on Lie group manifolds by making innovation updates in the tangent space. In all these approaches, however, the estimation problem was limited to the system kinematics with an assumption that the proprioceptive sensor, IMU, provides velocity measurements. 

Pertinent to observers which include motion dynamics, \cite{Salcudean} developed a globally convergent angular velocity observer using only orientation measurements, and used the natural energy function on the momentum as a Lyapunov candidate which resulted in a quadratic stable internal observer. \cite{Bras_noDoppler} and \cite{Bras_doppler} developed a nonlinear observer which estimated pose and a velocity and further demonstrated stability. Both these observers, however, used pose and velocity measurements. For non-cooperative targets, \cite{aghilli_EKF} developed a computationally efficient discrete EKF but the design did not exploit symmetries in motion and it is also not trivial to derive the region of attraction in an EKF. In contrast to \cite{Salcudean}, we propose an alternative approach to compute a vector difference using a push-forward vector operation. Furthermore, since the subject of this paper is concerned with a non-cooperative Target and velocity measurement is unavailable, the well-formalized theory of autonomous error dynamics in \cite{Bonnabel2}, \cite{MAHONY2013617} and \cite{Mahony} cannot be used directly and additionally, the observers in \cite{Bras_noDoppler} and \cite{Bras_doppler} are not applicable. Therefore, in this paper, we address the problem of estimating the inertial pose and body velocity of a non-cooperative free-floating Target using only the relative noisy pose measurements. 

The contributions of this paper are threefold. Firstly, we propose an observer which extends the existing concepts of kinematics symmetry in Lie group observer theory and uses additional properties of rigid body motion dynamics. To this end, we formalize the estimation problem on Lie groups with different state and measurement spaces and derive a novel left-invariant error formulation which narrows down observer analysis to the state-space error. The kinematics and dynamics part of the observer are designed such that it also acts as a predictor. Secondly, through Lyapunov analysis, we show that the observer error dynamics has almost-global uniform asymptotic stability. Through a reformulation of dynamics equations and exploitation of the skew-symmetric property in rigid body motion, we are able to simplify the stability analysis. Finally, we validate the proposed observer with $50$ Monte-Carlo simulations and experimental validation. 

The note is organized as follows. In sec~\ref{sec_mec_se3}, the underlying concepts of mechanical system modeling in $\text{SE}(3)$, which are relevant to this text, are provided. Following this, in sec.~\ref{sec_prob_form}, the problem is formalized on $\text{SE}(3)$ Lie group with a general measurement model. In sec.~\ref{sec_obs_des}, the proposed observer equations are derived and stability is proved for the observer parameters. This is followed by validation of the proposed method using robust Monte-Carlo simulations and experiments on OOS-SIM \cite{dlr97179} (see Fig.~\ref{fig_rigid}) in sec.~\ref{sec_results} and sec.~\ref{sec_exp} respectively. Finally, the conclusions and future scope of work are laid out in sec. \ref{sec_conc}. All the Lemmas that are used in the paper have been provided in the Appendix.

\section{Mechanical Systems on $\text{SE}(3)$ group} \label{sec_mec_se3}
Fig. \ref{fig_rigid} is a representative scenario for Target tracking using a manipulator-equipped spacecraft wherein, $\{\mathcal I\}$, $\{\mathcal T\}$, $\{\mathcal G\}$, and $\{\mathcal{E}\}$ indicate the inertial, Target center-of-mass (CoM), the grasping, and the end-effector-mounted camera frames respectively. Before introducing the kinematics and dynamics, we describe the notation concerning mechanical systems on $\text{SE}(3)$ which is used in this paper.
\subsection{Notations and definitions}
The pose (configuration) of a rigid body in space is given as a matrix Lie group representation called $\text{SE}(3)$ and is written as $g = \begin{bmatrix}R & p \\ 0 & 1 \end{bmatrix} \equiv (R,p)$, whose identity is $\mathbb{I}_{4,4}$, where $\mathbb{I}_{k,k}$ is a square identity matrix of dimension $k$. A pose between two frames is represented with the subscript of the lowercase letters of both frames. For instance, the pose of $\{\mathcal{G}\}$ relative to $\{\mathcal{T}\}$ is $g_{tg}$. The tangent space of a given pose $g \in \text{SE}(3)$ at identity is the velocity field (Lie Algebra) of the pose and is also a matrix group $\mathfrak{se}(3)$ which may be expressed in either the body ($V^b$) or a spatial ($V^s$) frame. In the following text, all velocity quantities are body velocities. Analogously, the cotangent space at identity, denoted as $\mathfrak{se}(3)^*$, is the space of generalized forces. An $\mathfrak{se}(3)$ velocity is given as $\begin{bmatrix}\omega_{\times} & v\\ 0 & 0 \end{bmatrix}$, where $(.)_{\times}$ indicates the skew-symmetric matrix for the vector and, $\omega$ and $v$ are angular and linear velocities respectively. $\mathfrak{se}(3)$ is isomorphic to $\mathbb{R}^6$, $[.]^\wedge:\mathbb{R}^6 \rightarrow \mathfrak{se}(3)$, $[.]^\vee:\mathfrak{se}(3) \rightarrow \mathbb{R}^6$ and in $\mathbb{R}^6$-form, is written as $V = \begin{bmatrix} \omega^T & v^T \end{bmatrix}^T$. Since $\mathfrak{se}(3)$ and corresponding $\mathbb{R}^6$ isomorphisms refer to different notations of the same quantity, they are used interchangeably for simplicity of notation in this paper. Poses and velocities with one subscript indicate that they are referenced relative to $\{I\}$, for instance, $\{g_t, ~V_t^b\}$ are pose and velocity of the Target CoM relative to $\{I\}$. 
\begin{definition}$\text{SE}(3)$ \emph{pose reconstruction formula:} \label{def_recons}
For a pose $g \in \text{SE}(3)$ and body velocity, $V^b$, $\dot{g} = g[V^b]^\wedge$. The superscript $b$ denotes body $\mathfrak{se}(3)$ velocity. 
\end{definition}

The Adjoint action, $\Ad:\mathfrak{se}(3) \rightarrow \mathfrak{se}(3)$, of a $\text{SE}(3)$ pose, $g$, transforms velocities between spatial and body frames as $V^s = Ad_{g}V^b$ where $\Ad_g = \begin{bmatrix} R & 0 \\ p_\times R & R \end{bmatrix}$, see \cite{SE3Control}. There also exists an adjoint map of the Lie Algebra onto itself, $\ad:\mathfrak{se}(3)\rightarrow\mathfrak{se}(3)$ which is the differential of the $\Ad$ map, $\ad_{V} = \begin{bmatrix}\omega_\times & 0\\ v_\times & \omega_\times \end{bmatrix}$ for $V \in \mathfrak{se}(3)$. The codajoint action $\ad^*:\mathfrak{se}(3)^* \rightarrow \mathfrak{se}(3)^*$ is defined as $\ad_{X}^* = \ad_{X}^T$. The $\text{SE(3)}$ Lie group and its algebra are endowed with a local (almost global) diffeomorphism map $\log:\text{SE(3)} \rightarrow \mathfrak{se}(3)$ and has been defined explicitly in Lemma \ref{lem_log} in the Appendix. 

The following two group operations are pointed out and will be used in sec.~\ref{sec_prob_form} to derive the measurement model.
\begin{definition}\emph{Lie group action:} \label{def_Lie_group}
     A Lie group action of a pose $g \in \text{SE}(3)$ on another group element $h \in \text{SE}(3)$, is a left and/or right translation operation(s), given as $\mathcal{L}_{g},~\mathcal{R}_{g}:\text{SE}(3)\rightarrow \text{SE}(3)$, $\mathcal{L}_{g}(h) = gh$, and $\mathcal{R}_g(h)= hg$, respectively.
\end{definition}
\begin{definition}\emph{Lie algebra Automorphism:}  \label{def_auto}
    Given $g,h \in \text{SE}(3)$ and a Lie group operation  $\Psi_g:\text{SE}(3) \rightarrow \text{SE}(3)$ such that $\Psi_g(h) = ghg^{-1}$, if $X = \log(h)$, then $Y = \log(\Psi_g(h)) = \Ad_gX$, for $X,Y \in \mathfrak{se}(3)$. In other words, if $\Psi_g$ is a group operation, $\Ad_g$ is its corresponding Lie algebra transformation.
\end{definition}

\subsection{Kinematics and Dynamics}
The Target (right) in Fig.~\ref{fig_rigid} is assumed to be a rigid body with fixed inertia and its configuration space is the pose $g_t \in \text{SE}(3)$ of $\{\mathcal{T}\}$. For a rigid body with inertia $\Lambda:\mathfrak{se}(3) \rightarrow \mathfrak{se}(3)^*$, the \emph{Euler-Poincar\'e} \cite[th. 6.1, iii]{Bloch1996} equation of motion is,
\begin{equation} \label{eq_lagrangian}
 \frac{d}{dt}\Lambda V^b = \ad_{V^b}^*\Lambda V^b   + f
\end{equation}
where $V^b \in \mathfrak{se}(3)$ and $f \in \mathfrak{se}(3)^*$.
Applying \eqref{eq_lagrangian} to a free-floating Target with $f = 0_{6,1}$, we get the equation of motion for inertia, $\Lambda_t = \begin{bmatrix}I_t & 0_{3,3}\\0_{3,3} & m_t\mathbb{I}_{3,3} \end{bmatrix}$, $V_t^b \in \mathfrak{se}(3)$ as,
\begin{numcases}{\text{Dynamics}}
    \Lambda_t\dot{V}_t^b = \ad_{V_t^b}^*\Lambda_tV_t^b.  \label{eq_target_dyn}
\end{numcases}
The kinematics for the Target are given by Def.~\ref{def_recons} as follows,
\begin{numcases}{\text{Kinematics}}
\dot{g_t} = g_t[V_t^b]^\wedge, ~g_t \equiv (R_t,r_t) \label{eq_target_kin}
\end{numcases}

In the analysis that follows, we use the following definitions. The set of singular values for $A \in \mathbb{R}^{n\times n}$ are given as, $\sigma(A) = \sqrt{\lambda(A^TA)}$, where $\lambda$ is the set of eigenvalues of $A$. $\underline{\sigma}(A) = \text{min}(\sigma)$ and $\overline{\sigma}(A) = \text{max}(\sigma)$ refer to the lowest and highest singular value of $A$ respectively. Additionally, the $l_2$ operator norm, $||A|| = \overline{\sigma}(A)$. For a vector $x \in \mathbb{R}^n$, such that $||\dot{x}|| \leq c, ~c>0$ implies $x \in \mathcal{C}^1$, which means that $x$ is continuous and $||{x}|| \leq b, ~b>0$ implies boundedness. $\because/\therefore$ refer to standard logical substitutes for because/therefore.  $\langle.,.\rangle$ refers to the inner product of the two arguments. $\mathbb{I}_{k,1}$ indicates a vector of ones.
\section{Problem formulation} \label{sec_prob_form}
In this section, the problem of estimating Target states, $\{g_t,V_t^b\}$ is formalized. This problem is abstractly illustrated in Fig. \ref{fig_2}. Illustrated on the left are the Lie group configuration-space trajectories of the true state ($g_t$, solid) and observer's estimate ($\hat{g}_t$, dashed) respectively. $\eta$ denotes the state estimation error between $g_t(t)$ and $\hat{g}_t(t)$. On the right are the measurement-space trajectories for the actual measurement (${g}$, solid) and estimated measurement ($\hat{g}$, dashed). $\Delta h$ denotes the measurement error which is the residual between $\hat{g}$ and $g$. Between configuration-space and measurement-space, there is a transformation, $h(.)$, obtained through Lie group actions (see Def.~\ref{def_Lie_group}). This means that, given $g_t, \hat{g}_t$, we can obtain $g = h(g_t)$ and $\hat{g} = h(\hat{g}_t)$ in the measurement space.  The poses, $g_t$ and $\hat{g}_t$ are associated to their Lie algebra $V_t^b,~\hat{V}_t^b \in \mathfrak{se}(3)$ respectively. The estimation problem in this paper is to use the measurement $g = h(g_t)$ to reconstruct the states which include the pose ($g_t$) and its Lie algebra ($V_t^b$). For this, we seek a transformation of errors from measurement-space to the configuration-space so that we can design the observer only in terms of the latter for simplicity.

\subsection{Configuration-space error}
\begin{figure}[t]
  \centering  	
    \psfrag{Gt1}[cc][cc][\FontFigS]{{\color{black}$g,\hat{g}\in \text{SE}(3)$}}
    \psfrag{Vt}[cc][cc][\FontFigM]{{\color{black}$V_t$}}
    \psfrag{Vthat}[cc][cc][\FontFigM]{{\color{black}$\hat{V}_t$}}
    \psfrag{gthat}[cc][cc][\FontFigM]{{\color{black}$\hat{g}_t$}}
    \psfrag{gt}[cc][cc][\FontFigM]{{\color{black}$g_t$}}
    \psfrag{eta}[cc][cc][\FontFigM]{{\color{black}$\eta$}}
    \psfrag{hgt}[cc][cc][\FontFigM]{{\color{black}$h(g_t)$}}
    \psfrag{hgh}[cc][cc][\FontFigM]{{\color{black}$h(\hat{g}_t)$}}
    \psfrag{Gt,Gth}[cc][cc][\FontFigS]{{\color{black}$g_t,\hat{g}_t \in \text{SE}(3)$}}
    \psfrag{delh}[cc][cc][\FontFigM]{{\color{black}$\Delta h$}}
    \includegraphics[width=0.40\textwidth]{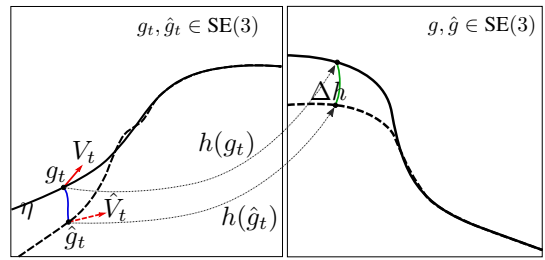}
    \caption{Manifold portrait showing estimation on Lie groups: Trajectory evolution of a rigid body and its observer with the same geometric structure.} \label{fig_2}
\end{figure}

First, we define a configuration-space error and a corresponding Lie algebra error as follows.
In Fig.~\ref{fig_2}, the estimation error in configuration-space, $\eta \in \text{SE(3)}$,  is defined as, $\eta = \hat{g}_t^{-1}g_t$. Note that $\eta$ is a left-invariant error formulation \cite[eq.~6]{Mahony}. Using the logarithm mapping defined in Lemma \ref{lem_log}, we obtain an error term as $\log(\eta) = [\epsilon]^\wedge \in \mathfrak{se}(3)$ and $\epsilon = \begin{bmatrix} \psi_t^T & q_t^T \end{bmatrix}^T$, where $\psi_t$ and $q_t$ are the orientation and translational errors respectively.
\subsection{Measurement-space error}
In this subsection, by using Lie group theory concepts, we establish a relationship between configuration-space errors ($\eta,\epsilon$) and a similar measurement-space error ($\Delta h,e$) so that in the rest of the text we can preclude measurement-space for simplicity.

To this end, we describe any general measurement model for $g$ as a composite Lie group action (Def.~\ref{def_Lie_group}) which may contain both $\mathcal{R}_{g}$ and $\mathcal{L}_{g}$ actions, given by a transformation $h(.):\text{SE}(3) \rightarrow \text{SE}(3)$. Let us assume that there exist both, left and right actions $g_l,~g_r \in \text{SE(3)}$ on the state $g_t$ such that $g = h(g_t) = g_lg_tg_r$. Using this, a left-invariant measurement pose error $\Delta h \in \text{SE(3)}$ is defined as $\Delta h = h(\hat{g}_t)^{-1}h({g}_t)$. A Lie algebra error $e \in \mathfrak{se}(3)$ is obtained using Lemma \ref{lem_log}, such that $e = \log(\Delta h)$. In the next step, we derive explicit forms of the relationship between errors.
Through rearrangement, we get,
\begin{equation} \label{eq_err_trafo1}
\eta = g_r\Delta h g_r^{-1}
\end{equation} 
The expression in \eqref{eq_err_trafo1} is exactly the $\Psi_{g_r}(\Delta h)$ operation (see Def.~\ref{def_auto}) of the action of $g_r$ contained in $h$. Hence, $\log(\Psi_{g_r}(\Delta h))$ undergoes an $\Ad_g$ transformation.
\begin{equation} \label{eq_erro_trafo}
    \epsilon = \Ad_{g_r}e
\end{equation}
From Fig.~\ref{fig_rigid}, the camera measurement pose is $g$ while the rigid body state is $g_t$. In this specific case, from kinematics we obtain, $g_r = g_{tg}$ and $g_l = g_e^{-1}$. Applying the aforementioned transformations in \eqref{eq_err_trafo1} and \eqref{eq_erro_trafo}, we get, 
\begin{equation}  \label{eq_err_trafO_fig1}
\eta = g_{gt}\Delta h g_{gt}^{-1}, ~\epsilon = \Ad_{g_{tg}}[\log(\Delta h)]^\vee, ~\Delta h = (\hat{g}^{-1}g)
\end{equation}
where $\hat{g} = g_e^{-1}\hat{g}_tg_{gt}$.
It can be seen that using \eqref{eq_err_trafO_fig1}, the measurement-space errors have been transformed to the configuration-space for the case in Fig.~\ref{fig_rigid}. Hence, in the rest of the paper, we perform analysis only with respect to the configuration-space errors $\eta,\epsilon$ which are obtained using \eqref{eq_err_trafO_fig1}.
\section{Observer design} \label{sec_obs_des}
\subsection{Kinematics observer}
Let us consider the kinematic part of the observer with the same geometric structure as \eqref{eq_target_kin} with $\hat{(.)}$ for corresponding estimates. This is obtained by applying Def.~\ref{def_recons} with an error injection term as follows,
\begin{equation}
\begin{aligned} 
    \dot{\hat{g}}_t &= \hat{g}_t[\hat{V}_t^b + \Ad_\eta K_1\epsilon]^\wedge
\end{aligned} \label{eq_obs_eqns}
\end{equation}
where, $K_1:\mathfrak{se}(3) \rightarrow \mathfrak{se}(3)$ is the observer kinematic gain which is determined through stability analysis in sec.~\ref{sec_obs_des}.

\subsection{Error kinematics}
In this section, the observer error kinematics are derived. The observer velocity error, $V_e^b$, is defined as $V_e^b = \begin{bmatrix} \omega_e^{b^T} & v_e^{b^T}\end{bmatrix}^T = V_t^b -  \Ad_{\eta^{-1}}\hat{V}_t^b$. Taking the time derivative of the pose error, $\eta = \hat{g}_t^{-1}g_t$, we get,
\begin{align}
    \begin{split} \label{eq_kin_pose}
    & \dot{\eta}  =  -\hat{g}_t^{-1} \dot{\hat{g}}_t\hat{g}_t^{-1} + \hat{g}_t^{-1}\dot{g}_t \\
    &\Rightarrow \dot{\eta} = \eta[V_t^b - \Ad_{(\eta^{-1})}\hat{V}_t^b -K_1\epsilon]^\wedge \\
    &\Rightarrow [{\eta^{-1}\dot{\eta}}]^\vee = \underbrace{V_t^b - \Ad_{(\eta^{-1})}\hat{V}_t^b}_{V_e^b} -K_1\epsilon 
    \end{split}\\
    \begin{split} \label{eq_kin_eq_1}
    & \Rightarrow  \dot{\epsilon}  = \mathcal{B}_r(\epsilon)(V_e^b -K_1\epsilon) \quad (\because \text{theorem \eqref{th_1}, Appendix})
    \end{split}
\end{align}
In parlance with the nomenclature presented in \cite[\S 4]{SE3Control}, the use of the $\log$ mapping for the error $\epsilon$ implies, the proposed observer is the \textit{Logarithm feedback} variant.

\subsection{Velocity error dynamics} \label{sec_dyn_obser_discs}
Before describing the equations of the dynamics observer, we first motivate the formulation by pointing out the following concept. For stability analysis, we need the velocity error dynamics from the equations of motion. In order to obtain this, the Target velocity ($V_t^b$) and observer velocity ($\hat{V}_t^b$) have to be compared as a valid vector operation. For a pose-error $\eta:\text{SE}(3) \rightarrow \text{SE}(3)$, $\Ad_{\eta^{-1}}:\mathfrak{se}(3) \rightarrow \mathfrak{se}(3)$ is the push-forward which transforms the velocity $\hat{V}_t^b$ to the actual Target body frame $\{\mathcal{T}\}$ as $\Ad_{\eta^{-1}}\hat{V}_t^b$. Hence, with the use of the push-forward and referring both velocities on the same point in $\text{SE}(3)$, we obtain a correct vector comparison between transformed observer velocity, $\Ad_{\eta^{-1}}\hat{V}_t^b$ and Target velocity, $V_t^b$. Note that, this is evident in the velocity error in \eqref{eq_kin_pose} which takes the form, $V_e^b = V_t^b - \Ad_{\eta^{-1}}\hat{V}_t^b$.

Following the discussion above, we compute the velocity error dynamics by taking the time-derivative of $V_e^b$. To this end, the time-derivative of $\Ad_{\eta^{-1}}\hat{V}_t^b$ is computed using \cite[Lemma $15$]{SE3Control} and \eqref{eq_kin_pose} and error dynamics are written as,
\begin{equation}
\begin{split} \label{eq_vel_err_dyn1}
    \frac{d}{dt}V_e^b &= \frac{d}{dt}({V}_t^b - \Ad_{(\eta^{-1})}{\hat{V}}_t^b)\\  =& \dot{V}_t^b - \big(\Ad_{(\eta^{-1})}\dot{\hat{V}}_t^b - \ad_{(V_e^b - K_1\epsilon)}\Ad_{\eta^{-1}}\hat{V}_t^b \big) 
    \end{split}
\end{equation}
\text{Substituting for} $\Ad_{(\eta^{-1})}{\hat{V}}_t^b = V_t^b - V_e^b$ and using the properties, $\ad_xx = 0$ and $\ad_xy = -\ad_yx$ in \eqref{eq_vel_err_dyn1}, we get
\begin{equation}
\label{eq_vel_err_dyn} 
    \frac{d}{dt}V_e^b = \dot{V}_t^b - \big(\Ad_{(\eta^{-1})}\dot{\hat{V}}_t^b + \ad_{V_t^b}V_e^b + \ad_{K_1\epsilon}\Ad_{\eta^{-1}}\hat{V}_t^b \big)
\end{equation}
Therefore, the observer design is now limited to determining $\Ad_{(\eta^{-1})}\dot{\hat{V}}_t^b$ such that the observer exhibits a stability property.

\subsection{Dynamics observer}
The observer dynamic equations of motion are proposed with a geometric structure similar to \eqref{eq_lagrangian} and velocity $\Ad_{\eta^{-1}}\hat{V}_t^b$ as, 
\begin{align}
    \begin{split}\label{eq_dyna_obs}
    \Lambda_t\Ad_{(\eta^{-1})}\dot{\hat{V}}_t^b =& \ad_{(\Ad_{\eta^{-1}}\hat{V}_t^b)}^*\Lambda_t\Ad_{(\eta^{-1})}\hat{V}_t^b+f_o \\ & - \Lambda_t\ad_{K_1\epsilon}\Ad_{\eta^{-1}}\hat{V}_t^b
    \end{split}
\end{align}
where $f_o \in \mathfrak{se}(3)^*$ is a design input force which is determined by Lyapunov stability analysis. It is also worth pointing out that the resultant system consisting of \eqref{eq_obs_eqns} and \eqref{eq_dyna_obs} is an internal observer (see \cite{Bonnabel2}, \cite{MAHONY2013617}), which means that the time-evolution of the system in absence of measurement is like that of a rigid body and hence the observer can be used as a predictor.

Substituting \eqref{eq_dyna_obs} in \eqref{eq_vel_err_dyn}, we get,
\begin{equation}
    \begin{split}
        \frac{d}{dt}V_e^b &= \Lambda_t^{-1}\big(\ad_{V_t^b}^*\Lambda_tV_t^b - \ad_{(\Ad_{\eta^{-1}}\hat{V}_t^b)}^*\Lambda_t\Ad_{(\eta^{-1})}\hat{V}_t^b \\& - \Lambda_t\ad_{V_t^b}V_e^b -f_o)
    \end{split}
\end{equation}
Applying Lemma \ref{lem_vel_err} to the first two terms in R.H.S, we obtain,
\begin{align} \label{eq_vel_err_dyn_main}
\begin{split}
\frac{d}{dt}V_e^b =& \Lambda_t^{-1}\big(\underbrace{(\ad_{V_t^b}^*\Lambda_t + \ad_{\Lambda_t V_t^b}^\sim - \ad_{V_e^b}^*\Lambda_t -\Lambda_t\ad_{V_t^b})}_{\mathcal{C}(V_t^b,V_e^b)}V_e^b  \\ &- f_0\big)
\end{split}
\end{align}
The observer error dynamics can be written together for kinematics in \eqref{eq_kin_eq_1} and dynamics in \eqref{eq_vel_err_dyn_main} as,
\begin{align}
    \begin{split}\label{eq_obs_err_dyn}
    \frac{d}{dt}\begin{bmatrix}{\epsilon} \\ V_e^b\end{bmatrix} =& \underbrace{\begin{bmatrix}-\mathcal{B}_r(\epsilon)K_1 & \mathcal{B}_r(\epsilon)\\ 0_{6,6} & \Lambda_t^{-1}\mathcal{C}(V_t^b,V_e^b) \end{bmatrix}}_{{A}}\begin{bmatrix} \epsilon \\ V_e^b \end{bmatrix} \\ & - \underbrace{\begin{bmatrix}0_{6,6} \\ \Lambda_t^{-1} \end{bmatrix}}_{B}f_0
    \end{split}
\end{align}

In the following, $K_1$ and $f_0$ are determined using Lyapunov stability analysis for the proposed nonlinear observer.
 
\begin{theorem} \emph{Main result}: \label{th_main_stab}
For a free-floating rigid body whose motion equations are given by \eqref{eq_target_kin} and \eqref{eq_target_dyn}, and an observer given by \eqref{eq_obs_eqns} and \eqref{eq_dyna_obs}, the observer error, $x = \begin{bmatrix}\epsilon^T & V_e^T\end{bmatrix} \in \mathbb{R}^{12}$ is almost globally uniformly asymptotically stable about the origin for design parameters $P_1,P_2$, and observer kinematic gain, $K_1$ and dynamic input $f_0$ such that,
\begin{enumerate}
    \item $K_1 = k_1\mathbb{I}_{6,6} = P_1^{-1} =\frac{1}{p_{1}}\mathbb{I}_{6,6}, ~p_{1},k_{1}> 0$
    \item $P_2 = \text{diag}(p_2), ~p_2  = [p_{21}\mathbb{I}_{3,1}^T,~p_{22}\mathbb{I}_{3,1}^T]^T,~p_{21},p_{22} > 0$
    \item $f_0 = p_{1}P_2^{-1}\mathcal{B}_r(\epsilon)^T\epsilon$
    \item $p_{1}> \frac{p_{22}\overline{\sigma}(\Lambda_t)||\omega_e^b(0)||^2}{\pi^2 -||\psi_t(0)||^2} $ \label{it_almost_global}
\end{enumerate}
Proof:
The proof is split into two parts. In the first part, uniform asymptotic stability is proved and in the latter part, the constraint on the matrix $P$ is determined so that the rotational singularity in the $\log$ map is not encountered along trajectories. The latter part refers to the item \ref{it_almost_global} in theorem \ref{th_main_stab} and ensures almost-global stability of the observer error system in \eqref{eq_obs_err_dyn}.
Choosing the Lyapunov candidate as $W = \frac{1}{2}x^TP x$, where $P =\begin{bmatrix}P_1 & 0_{6,6}\\ 0_{6,6} & P_2\Lambda_t\end{bmatrix}$ such that $P = P^T$ for an open connected region $x(0) \in \Omega \subset \mathbb{R}^{12}$, and there exist bounds as,
\begin{equation} \label{eq_lyap_bounds}
    \frac{1}{2}\underbrace{\underline{\sigma}({P})||x||^2}_{\underline{\alpha}(||x||)} \leq W \leq \frac{1}{2} \underbrace{\overline{\sigma}({P})||x||^2}_{\overline{\alpha}(||x||)}
\end{equation}
Taking time derivative along trajectories and using observer error dynamics in \eqref{eq_obs_err_dyn}, we get,
\begin{align}
    \begin{split} \label{eq_reference_cancel}
        \dot{W} =& x^TPAx -x^TPBf_o\\ 
         = &x^T\begin{bmatrix} - P_1\mathcal{B}_r(\epsilon)K_1 & P_1\mathcal{B}_r(\epsilon) \\0_{6,6} &  P_2\mathcal{C}(V_t^b,V_e^b)  \end{bmatrix}x\\ & - x^T\begin{bmatrix}0_{6,6}\\ P_2\end{bmatrix} f_o \\
    \end{split}
\end{align}
In the following, constraints are imposed on the design parameters in order to simplify \eqref{eq_reference_cancel}. We choose, $K_1 = k_1\mathbb{I}_{6,6}= \frac{1}{p_{1}}\mathbb{I}_{6,6}$ so that we can apply Lemma \ref{lem_br_eps} to the block matrix $(1,1)$ position in the first term. The Lemma \ref{lem_innerProd} is applied to the block matrix position $(2,2)$ in the first term so that the term with $\mathcal{C}(V_t^b,V_e^b)$ vanishes. Furthermore, we set the input $f_o=p_{1}P_2^{-1}\mathcal{B}_r(\epsilon)^T\epsilon$ which leads to a cancellation of coupled terms, ($(2,1)$ and $(1,2)$) that follow in eq.\eqref{eq_cancel_br}. Hence, we get,
\begin{align} 
      \dot{W} = &x^T\begin{bmatrix}- k_1p_{1}\mathbb{I}_{6,6} & p_{1}\mathcal{B}_r(\epsilon)\\-p_{1}\mathcal{B}_r(\epsilon)^T & 0_{6,6} \end{bmatrix}x \label{eq_cancel_br}\\
      = & -k_1p_{1}||\epsilon||^2 \leq 0 \label{eq_lyap_nsd}
\end{align}
Hence, from \eqref{eq_lyap_nsd}, we first conclude that the observer error dynamics in \eqref{eq_obs_err_dyn} is uniformly stable. 
In order to prove uniform asymptotic stability of the state $x$, we use Matrosov's theorem (see \cite{Paden} for application) from theorem \ref{th_matros} (in Appendix). We choose an auxiliary function, $\mathcal{W} = x^T\mathcal{P}x$, where $\mathcal{P} =\begin{bmatrix}{P}_{1} & 0_{6,6}  \\ -\Lambda_t &{P}_{2}\Lambda_t \end{bmatrix}$. 

Using Lemma \ref{lem_bounded_terms} which guarantees a bounded observer error $x$, we deduce that 
\begin{equation} \label{eq_W_bounded_corr}
 \underline{\beta}||x||^2\leq  |\mathcal{W}| \leq \overline{\beta}||x||^2
\end{equation}
where $\underline{\beta}, \overline{\beta} > 0$. Hence, $|\mathcal{W}|$ is bounded.

Taking time derivative of $\mathcal{W}$ along trajectories, setting $P_2 = \text{diag}(p_2)$ and $f_o$ as defined in theorem \ref{th_main_stab}, we obtain,
\begin{equation}
    \begin{split}\label{eq_mathW_td}
         \dot{\mathcal{W}} &=x^T(\underbrace{\mathcal{P}A + A^T\mathcal{P}}_{Q_1})x - x^T\mathcal{P}Bf_o - f_o^TB^T\mathcal{P}x \\
        =& x^T{Q_1}x - 2p_{1}V_e^{b^T}\mathcal{B}_r(\epsilon)^T\epsilon + {p_{1}}\epsilon^TP_{2}^{-1}\mathcal{B}_r(\epsilon)^T\epsilon \\
        =& x^T({Q_1} - {Q_2})x \\
    \end{split}
\end{equation}
where $Q_2 = \begin{bmatrix}-{p_{1}}P_{2}^{-1}\mathcal{B}_r(\epsilon)^T & p_{1}\mathcal{B}_r(\epsilon)\\ p_{1}\mathcal{B}_r(\epsilon)^T & 0_{6,6} \end{bmatrix}$. Continuity and boundedness of $\dot{\mathcal{W}}$ follows from the conclusion in Lemma \ref{lem_bound_cont}, which was derived by systematically proving these two properties for all the terms in \eqref{eq_mathW_td}.

Furthermore, In the set $\{x \in \Omega|\dot{W} = 0\} \equiv \{x \in \Omega|~||\epsilon|| = 0\}$, applying limits to \eqref{eq_mathW_td}, we obtain,
\begin{equation}
    \begin{split}
        \Rightarrow &\lim_{\dot{W} \rightarrow 0} \dot{\mathcal{W}} =\lim_{||\epsilon|| \rightarrow 0}  \dot{\mathcal{W}}= -V_e^{b^T}\Big(\Lambda_t\mathcal{B}_r(0_{6,1})   \\&-(\mathcal{C}(V_t^b,V_e^b)^TP_{2}
         +P_{2}\mathcal{C}(V_t^b,V_e^b)\Big)V_e^b
    \end{split}
\end{equation}
Using the definition of $\mathcal{B}_r(\epsilon)$ from Lemma \ref{lem_exp_jacob}, $\mathcal{B}_r(0) = \mathbb{I}_{6,6}$. Employing the inner product property of Lemma \ref{lem_innerProd} (see \eqref{eq_sum_skew_sym}) to cancel out terms with $\mathcal{C}(V_t^b,V_e^b)$, we obtain,
\begin{equation} \label{eq_W_bounded}
    \begin{split}
    \lim_{\dot{W} \rightarrow 0} \dot{\mathcal{W}} &= -V_e^{b^T}\Lambda_tV_e^b \leq  - \underline{\sigma}(\Lambda_t)||V_e^b||^2\\
    \Rightarrow \lim_{\dot{W} \rightarrow 0} |\dot{\mathcal{W}}| & \geq \underline{\sigma}(\Lambda_t)||V_e^b||^2
    \end{split}
\end{equation}

We conclude therefore that $\mathcal{W}$ is bounded and sign-definite (negative for non-zero values of $||V_e^b||$) in the set $\{x \in \Omega|~||\epsilon|| = 0\}$. 

The conclusions from \eqref{eq_lyap_bounds}, \eqref{eq_lyap_nsd}, \eqref{eq_W_bounded_corr} match conditions $1,~2,~3$ in Matrosov's theorem. For the condition $4$, the two conditions in Lemma \ref{lem_matros} are satisfied by Lemma \ref{lem_bound_cont} and \eqref{eq_W_bounded}. Since, the error dynamics in \eqref{eq_obs_err_dyn} is bounded for $x \in \Omega$, from the Matrosov's theorem in Lemma \ref{lem_matros}, we conclude uniform asymptotic stability of the state $x$ about the origin.

Topological drawbacks preclude global stability in $\text{SE(3)}$ due to the ambiguity in rotation. Readers are referred to works by \cite{Bras_noDoppler}, \cite{Koditschek} and \cite{BULLO199917} where this problem is discussed and applied. In order to achieve, almost-global stability, a condition on the minimum $p_{1}$ is derived next. In this part of the analysis, all functions with a subfix $(.)_{\omega}$ denote the rotational component of the corresponding function.

Let us define a sublevel set, $\Omega_\omega = \{x \in \mathbb{R}^{12}: W_\omega < \frac{p_1\pi^2}{2}\}$, which is the state-space at the least value of $W$ at singularity ($||\psi|| = \pi,\omega_e^b = 0_{3,1}$). Since, we proved that $W$ is non-increasing and furthermore, $W_\omega, \dot{W}_\omega$, and choice of $P$ are decoupled from the translational part, $\Omega_\omega$ is a positive invariant set. If the upper bound of $W_{\omega}(t=0)$ for time $t$ is restricted within the aforesaid sublevel set $\Omega_\omega = \{x \in \mathbb{R}^{12}: W_\omega < \frac{p_1\pi^2}{2}\}$, it is sufficient to ensure that the observer trajectories never encounter the rotational singularity. This can be written as, 
\begin{equation}
    \begin{split}\label{eq_rot_bound_proof}
    &\frac{1}{2}\big(p_{1}||\psi_t(0)||^2 +p_{22}\overline{\sigma}(I_t)||\omega_e^b(0)||^2\big) < p_{1}\frac{\pi^2}{2}\\
    \Rightarrow & p_{1} > \frac{p_{22}\overline{\sigma}(I_t)||\omega_e^b(0)||^2}{\pi^2 - ||\psi(0)||^2}
    \end{split}
\end{equation}
\eqref{eq_rot_bound_proof} provides a sufficient condition to ensure that the observer error dynamics in \eqref{eq_obs_err_dyn} have almost-global uniform asymptotic stability. 
\QEDB
\end{theorem}

\section{Monte-Carlo Simulation results} \label{sec_results}
The proposed observer was implemented for estimating the states of an inactive tumbling satellite (ENVISAT, \cite{ESA_comrade}). In such a scenario, the motion states as well as the the inertia, $\Lambda_t$, are subject to uncertainties (detailed in Table \ref{tbl_details}). These uncertainties, together with the exteroceptive sensor (camera) noise, make estimation robustness a key criteria in determining practical use. The camera noise was simulated as a concentrated Gaussian \cite{Bourmaud2015} with $\nu = 1e^{-4}\mathbb{I}_{6,1}$ in the tangent space and the noisy measurement was obtained as $\tilde{g} = g\exp{[\nu]^\wedge}$ with sampling time $T = 0.1\text{[s]}$. In order to validate the robustness of the proposed observer, $50$ Monte-Carlo simulations were performed by varying the set $\{\Lambda_t,g_t(0),V_t(0)\}$ within the uncertainty bounds as specified in Table \ref{tbl_details}. In all the simulations, the observer was initialized only to zero initial conditions and the gains were computed from parameters $P_1$ and $P_2$ which are declared in Table \ref{tbl_details}. In the description below, $\eta \equiv (R(\theta_e),p_e)$ is used to show position and orientation errors.

\begin{figure}[!h]
    \centering  
    \psfrag{time}[cc][cc][\FontFigXS]{{\color{black}$t\text{[s]}$}}
    \psfrag{p1}[cc][cc][\FontFigXS]{{\color{black}$p_{tx}$}}
    \psfrag{p2}[cc][cc][\FontFigXS]{{\color{black}$p_{ty}$}}
    \psfrag{p3}[cc][cc][\FontFigXS]{{\color{black}$p_{tz}$}}
    \psfrag{p4}[cc][cc][\FontFigXS]{{\color{black}$\hat{p}_{tx}$}}
    \psfrag{p5}[cc][cc][\FontFigXS]{{\color{black}$\hat{p}_{ty}$}}
    \psfrag{p6}[cc][cc][\FontFigXS]{{\color{black}$\hat{p}_{tz}$}}
    \psfrag{v1}[cc][cc][\FontFigXS]{{\color{black}$v_{tx}$}}
    \psfrag{v2}[cc][cc][\FontFigXS]{{\color{black}$v_{ty}$}}
    \psfrag{v3}[cc][cc][\FontFigXS]{{\color{black}$v_{tz}$}}
    \psfrag{v4}[cc][cc][\FontFigXS]{{\color{black}$\hat{v}_{tx}$}}
    \psfrag{v5}[cc][cc][\FontFigXS]{{\color{black}$\hat{v}_{ty}$}}
    \psfrag{v6}[cc][cc][\FontFigXS]{{\color{black}$\hat{v}_{tz}$}}
    \psfrag{om1}[cc][cc][\FontFigXS]{{\color{black}$\omega_{tx}$}}
    \psfrag{om2}[cc][cc][\FontFigXS]{{\color{black}$\omega_{ty}$}}
    \psfrag{om3}[cc][cc][\FontFigXS]{{\color{black}$\omega_{tz}$}}
    \psfrag{om4}[cc][cc][\FontFigXS]{{\color{black}$\hat{\omega}_{tx}$}}
    \psfrag{om5}[cc][cc][\FontFigXS]{{\color{black}$\hat{\omega}_{ty}$}}
    \psfrag{om6}[cc][cc][\FontFigXS]{{\color{black}$\hat{\omega}_{tz}$}}
    \psfrag{q1}[cc][cc][\FontFigXS]{{\color{black}$q_{t0}$}}
    \psfrag{q2}[cc][cc][\FontFigXS]{{\color{black}$q_{tx}$}}
    \psfrag{q3}[cc][cc][\FontFigXS]{{\color{black}$q_{ty}$}}
    \psfrag{q4}[cc][cc][\FontFigXS]{{\color{black}$q_{tz}$}}
    \psfrag{q5}[cc][cc][\FontFigXS]{{\color{black}$\hat{q}_{t0}$}}
    \psfrag{q6}[cc][cc][\FontFigXS]{{\color{black}$\hat{q}_{tx}$}}
    \psfrag{q7}[cc][cc][\FontFigXS]{{\color{black}$\hat{q}_{ty}$}}
    \psfrag{q8}[cc][cc][\FontFigXS]{{\color{black}$\hat{q}_{tz}$}}
    
    \psfrag{vt}[cc][cc][\FontFigM]{{\color{black}$v_t$}}
    \psfrag{omt}[cc][cc][\FontFigM]{{\color{black}$\omega_t$}}
    \psfrag{pt}[cc][cc][\FontFigM]{{\color{black}$p_t$}}
    \psfrag{qt}[cc][cc][\FontFigM]{{\color{black}$q_t$}}
    \includegraphics[width=0.45\textwidth]{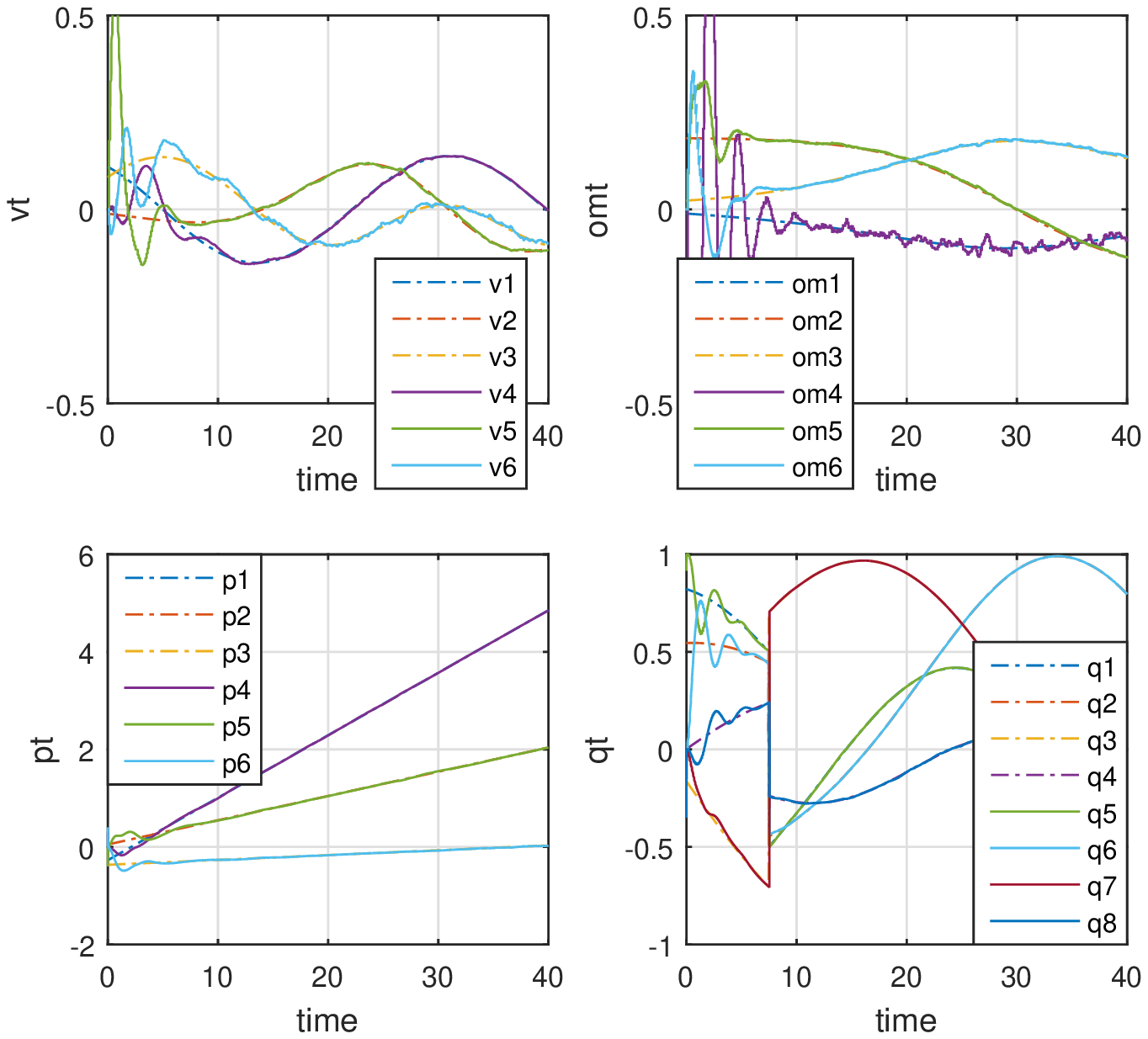}
    \caption{Convergence of estimates ($(\hat{.})$, solid) to the ground-truth ($(.)$, dashed) for the $50^{\text{th}}$ Monte-Carlo simulation.} \label{fig_state_conv}

    \centering  
    \psfrag{v}[cc][cc][\FontFigS]{{\color{black}$v_{ei}\text{[m/s]},~i=[x,y,z]$}}
    \psfrag{om}[cc][cc][\FontFigS]{{\color{black}$\omega_{ei}\text{[\degree/s]},~i=[x,y,z]$}}
    \psfrag{p}[cc][cc][\FontFigS]{{\color{black}${p}_{ei}\text{[m]},~i=[x,y,z]$}}
    \psfrag{q}[cc][cc][\FontFigS]{{\color{black}$\theta_{ei}[^{\circ}],~i=[x,y,z]$}}
    \psfrag{v1}[cc][cc][\FontFigM]{{\color{black}$n$}}
    \psfrag{v4}[cc][cc][\FontFigM]{{\color{black}$n$}}
    \psfrag{p1}[cc][cc][\FontFigM]{{\color{black}$n$}}
    \psfrag{q1}[cc][cc][\FontFigM]{{\color{black}$n$}}
    \psfrag{x}[cc][cc][\FontFigXS]{{\color{black}$x$}}
    \psfrag{y}[cc][cc][\FontFigXS]{{\color{black}$y$}}
    \psfrag{z}[cc][cc][\FontFigXS]{{\color{black}$z$}}
    \includegraphics[width=0.45\textwidth]{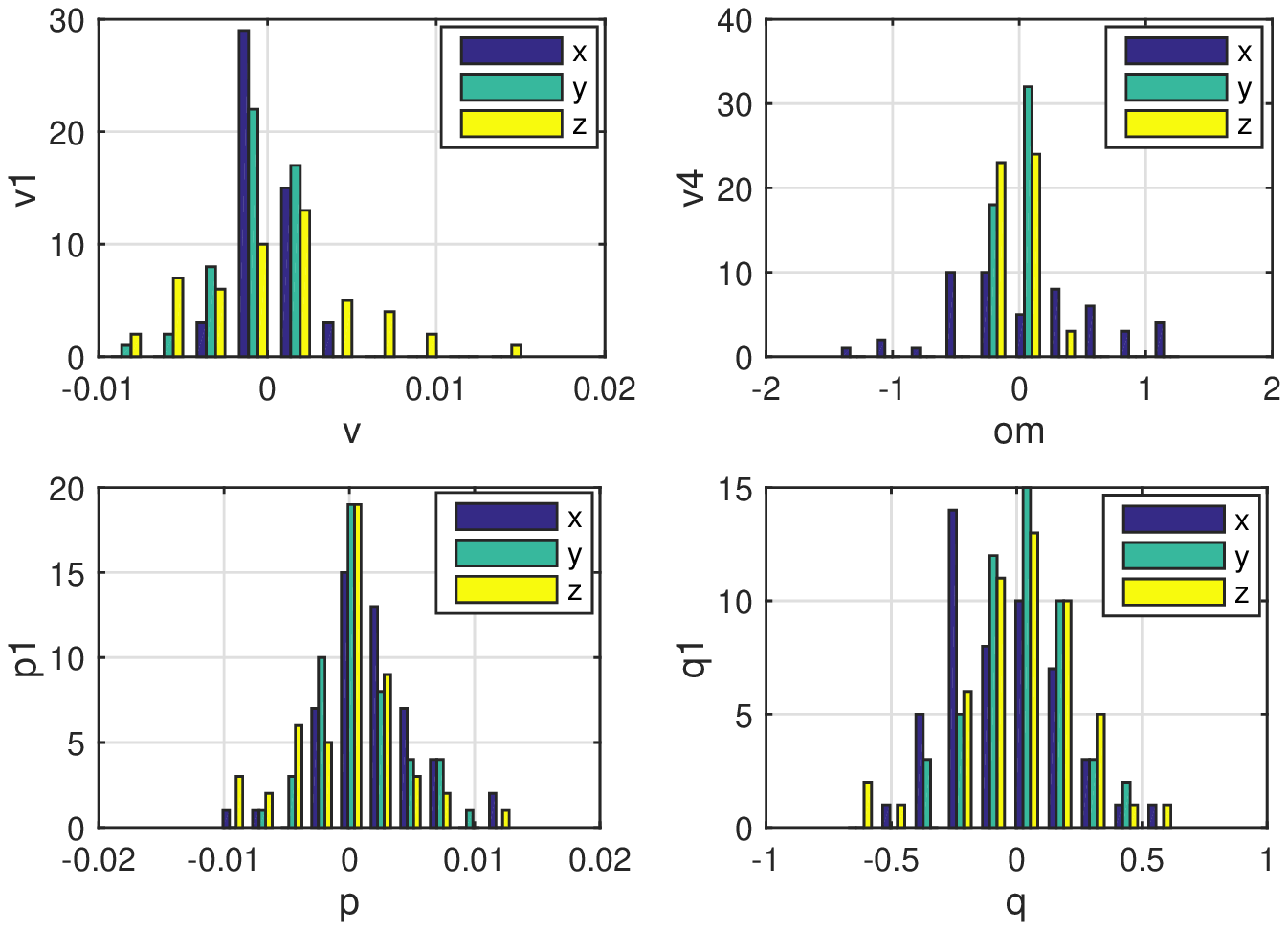}
    \caption{Histogram (component-wise) of errors in $50$ Monte-Carlo simulations for velocities(top row) and pose (bottom row) with specified uncertainties in Table \ref{tbl_details}.} \label{fig_hist}
\end{figure}

Fig.~\ref{fig_state_conv} shows the convergence of all motion states: the configuration pose $g_t$ and velocity $V_t$, for the $50^\text{th}$ simulation run. The velocity (top row), $V_t = \begin{bmatrix}\omega_t^T & v_t^T \end{bmatrix}^T$, of the Target is shown to converge after an initial transient period. The configuration pose, $g_t$, which is plotted as position and quaternion, converged to position-error norm, $||p_{e}|| = 0.0084 \text{[m]}$ and angular-error norm, $||\theta_{e}|| = 0.3\text{[\degree]}$. Additionally, in Fig.~\ref{fig_hist}, from the histogram of component-wise errors for velocities (top row) and pose (bottom row), we can infer that despite the introduced uncertainty, the observer converges to a bounded error. The results tabulated in Table \ref{tbl_summary} summarize these histogram results and it can be seen that the maximum error-norm of both, position and orientation, are $0.012\text{[m]}$ and $0.5271\text{[\degree]}$ respectively, which are better than the corresponding metrics for the noisy measurements ($0.0173 \text{[m]}$, $1\text{[\degree]}$ respectively). These results validate the design and additionally prove robustness of the observer which was designed using theorem \ref{th_main_stab}.

\begin{table}[t]
\caption{Simulation/Design of observer} \label{tbl_details}
\begin{center}
\begin{tabular}{| l | l | l | l |}
\hline
$\Lambda_t$ & $I_t(\text{[Kg.m}^2\text{]}) = \begin{bmatrix}17023.3 & 397.1 &  -2171.4\\397.1 &124825.7 & 344.2\\-2171.4 & 344.2 & 129112.2 \end{bmatrix}$\\& $\pm \begin{bmatrix}350 & 100 & 250\\100 & 3000  &150\\250 &  150 & 3000\end{bmatrix}$ \\ &$m_t(\text{[Kg]}) = 7827.867 \pm 78.27867$\\
\hline
$V_t^b(0)$ & $0_{6,1} \pm 0.0873\mathbb{I}_{6,1}(\text{[m/s]},\text{[rad./s]})$\\
\hline
${g}_t(0)$ & $(R(\pm 45 \text{[\degree]},\pm 45\text{[\degree]},\pm 45\text{[\degree]}, \pm 0.5\mathbb{I}_{3,1}[m])^*$\\
\hline 
$\hat{V}_t^b(0)$ & $\begin{bmatrix}0_{3,1}\text{[rad/s]}&0_{3,1}\text{[m/s]} \end{bmatrix}^T$ \\
\hline
$\hat{g}_t(0)$ & $(R(0,0,0), \begin{bmatrix}0 &0 & 0 \end{bmatrix}^T[m])^*$
\\
\hline
$P$& $p_{1} = 0.1042$, \\ &$P_2 =  1.0e^{-05}\text{diag}([0.1158\mathbb{I}_{3,1}^T,0.0124\mathbb{I}_{3,1}^T]^T)$\\
\hline
Th. \ref{th_main_stab}, \ref{it_almost_global}
&$p_{1} = 0.1042 > 1.3549e^{-05} =\frac{p_{22}\overline{\sigma}(\Lambda_t)||\omega_e^b(0)||^2}{\pi^2 -||\psi_t(0)||^2} $
\\

\hline
$K_1$ & $k_1 = 9.597$ \\
\hline
&$^*${\tiny{$\text{X-Y-Z}$-sequence Euler angle parameterization}}\\
\hline
\end{tabular}
\end{center}
\end{table}

\begin{table}[t]
\caption{Summary of Monte-Carlo simulations} \label{tbl_summary}
\begin{center}
\begin{tabular}{| l | l | l | l |}
\hline
& Mean $\pm$ std & Max($||.||$) & Min($||.||$)\\ \hline
$p_{e}\text{[m]}$ & $\begin{bmatrix}0.0017\\ 0.0007\\ -0.0003 \end{bmatrix}\pm \begin{bmatrix} 0.0038 \\   0.0033 \\    0.0043\end{bmatrix} $ & $0.0120$ & $8.4542e^{-04}$
\\
\hline
$\theta_{e}\text{[\degree]}$ & $\begin{bmatrix}-0.0531 \\   0.0248 \\   0.0057 \end{bmatrix} \pm \begin{bmatrix}0.2226  \\  0.1806  \\  0.2425 \end{bmatrix}$ & $0.5271$ &  $0.1082$\\
\hline
\end{tabular}
\end{center}
\end{table}
\section{Experimental validation} \label{sec_exp}
\begin{figure}[!h]
    \centering  
    \psfrag{v}[cc][cc][\FontFigS]{{\color{black}$v_{ei}\text{[m/s]},~i=[x,y,z]$}}
    \psfrag{om}[cc][cc][\FontFigS]{{\color{black}$\omega_{ei}\text{[\degree/s]},~i=[x,y,z]$}}
    \psfrag{time}[cc][cc][\FontFigS]{{\color{black}$t\text{[s]}$}}
    \psfrag{q}[cc][cc][\FontFigS]{{\color{black}$\theta_{ei}[^{\circ}],~i=[x,y,z]$}}
    \psfrag{om1}[cc][cc][\FontFigS]{{\color{black}$\hat{\omega}_2$}}
    \psfrag{om2}[cc][cc][\FontFigS]{{\color{black}$\hat{\omega}_3$}}
    \psfrag{om3}[cc][cc][\FontFigS]{{\color{black}$\hat{\omega}_4$}}
    \psfrag{om4}[cc][cc][\FontFigS]{{\color{black}${\omega}_2$}}
    \psfrag{om5}[cc][cc][\FontFigS]{{\color{black}${\omega}_3$}}
    \psfrag{om6}[cc][cc][\FontFigS]{{\color{black}${\omega}_4$}}
    \psfrag{ome}[cc][cc][\FontFigM]{{\color{black}$\omega_x\text{[rad./s]}$}}
    \includegraphics[width=0.45\textwidth]{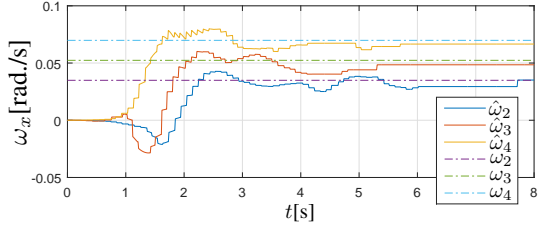}
    \caption{Comparison of estimates, $\hat{\omega}_x$, with ground truth, $\omega_x$, for an axially spinning Target simulated on the OOS-SIM facility (see Fig.~\ref{fig_rigid}) for $\omega_x = i,~i \in [2,3,4]\text{[\degree/s}]$ where pose measurements were obtained from a camera and image-processing system at $10\text{[Hz]}$.} \label{fig_exp}
\end{figure}

\begin{figure}[!h]
    \centering  
     \psfrag{time}[cc][cc][\FontFigS]{{\color{black}$t\text{[s]}$}}
      \psfrag{qt}[cc][cc][\FontFigM]{{\color{black}$q_t$}}
    \psfrag{q1}[cc][cc][\FontFigXS]{{\color{black}$q_0$}}
    \psfrag{q2}[cc][cc][\FontFigXS]{{\color{black}$q_x$}}
    \psfrag{q3}[cc][cc][\FontFigXS]{{\color{black}$q_y$}}
    \psfrag{q4}[cc][cc][\FontFigXS]{{\color{black}$q_z$}}
    \psfrag{q5}[cc][cc][\FontFigXS]{{\color{black}$\hat{q}_0$}}
    \psfrag{q6}[cc][cc][\FontFigXS]{{\color{black}$\hat{q}_x$}}
    \psfrag{q7}[cc][cc][\FontFigXS]{{\color{black}$\hat{q}_y$}}
    \psfrag{q8}[cc][cc][\FontFigXS]{{\color{black}$\hat{q}_z$}}
    \includegraphics[width=0.35\textwidth]{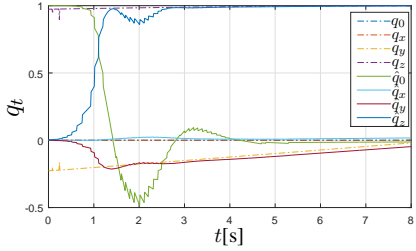}
    \caption{Convergence of orientation estimate ($\hat{q}_t \mapsto R_t(q_t)$), to ground truth, $q_t$, for an axially spinning Target at $\omega_x = 4\text{[\degree/s]}$.} \label{fig_exp2}
\end{figure}

The proposed observer was implemented at the OOS-SIM facility at DLR to estimate states of an axially spinning satellite. The satellite inertia was simulated using the facility's client dynamics with $\Lambda_t \equiv (m_t, I_t)$, $m_t = 341 \text{[Kg]}$ and $I_t = \text{diag}([400.1025,  262.9500, 264.9425]) \text{[Kg.m}^2\text{]}$. The satellite was spun about its dominant $x-$axis with $\omega_x = i,~i \in [2,3,4]\text{[\degree/s]}$ for $8\text{[s]}$ each time. The satellite was observed using a end-effector camera as shown in Fig.~\ref{fig_rigid} and an image-processing algorithm was used to provide pose-measurements to the observer at $10 \text{[Hz]}$. Firstly, the convergence of the estimated $\hat{\omega}_x$ towards ground truth $\omega_x$ is demonstrated in Fig.~\ref{fig_exp}. For practical purposes, a heuristic threshold-based outlier rejection scheme was implemented to avoid using unlikely pose-estimates from the image-processing algorithm. However, despite this, we observe that due to extremely noisy pose measurements, the estimation degrades and fluctuates about the true value. Furthermore, for the fastest case $\omega_x = 4\text{[\degree/s]}$, we demonstrate the convergence of the observer state $\hat{q}_t \mapsto R_t(\hat{q}_t)$ to the ground truth $q_t$ within $5\text{[s]}$ from $0$ initial conditions of the observer in Fig.~\ref{fig_exp2}. This concludes the experimental validation of the proposed observer.

\section{Conclusion} \label{sec_conc}
In this paper, a nonlinear rigid body observer was developed using the $\log$ coordinates of the $\text{SE(3)}$ pose error to estimate the current pose and the body velocity of a free-floating Target. The problem was first formalized for rigid body motion on the $\text{SE(3)}$ Lie group. Secondly, the kinematics and the dynamics equations were proposed such that the observer acted as an internal model that evolves in time as a rigid body and can be used as a predictor. By using a reformulation and skew-symmetric property of rigid body dynamics, the stability analysis of the error dynamics was simplified. The observer was proved to possess almost-global uniform asymptotic stability. Finally, through robust Monte-Carlo simulation results and experiments, the proposed method was verified and validated. As a future work, the observer shall be extended for a forced rigid-body motion with additional velocity and force measurements.

\appendix[Useful properties and lemmas]\label{app_1}
\begin{lemma} \label{lem_log}
Let $g \equiv (R,p) \in \text{SE(3)}$  be a group entity such that $\tr(R) \neq -1$, then
\begin{equation}
    \log(g) = \begin{bmatrix}\psi_\times  & A(\psi)^{-1}p\\0 & 1 \end{bmatrix} \Rightarrow  [\log(g)]^\vee = \begin{bmatrix}\psi\\q\end{bmatrix}
\end{equation}
where, 
$ A(\psi)^{-1} = (\mathbb{I}_{3,3} - \frac{1}{2}\psi_\times + (1-\alpha(|\psi|))\frac{\psi_\times^2}{|\psi|^2})$  with $\alpha(x) = \frac{x}{2}\cot(\frac{x}{2})$, $\psi = \log(R) = \frac{\phi}{2\sin\phi}(R-R^T)$, $\phi = \frac{1}{2}(\tr(R) - 1)$, $q = A(\psi)^{-1}p$.
\end{lemma}
\begin{lemma} (Differential of exponential)\cite{SE3Control} \label{th_diff_exp}
Let $g(t) \in \text{SE}(3)$ be a smooth curve, $X(t) = \log(g(t))$ and $x(t) = [X(t)]^\vee$  where $\log:\text{SE}(3)\rightarrow\mathfrak{se}(3)$ defines the logarithmic mapping, $V^b = g(t)^{-1}\dot{g}(t)$ is the body velocity,
\begin{align}
    \dot{x}(t) = \sum_{n=0}^\infty (-1)^n \frac{B_n}{n!}\ad_X^n[V^b]^\vee = \mathcal{B}_r(X)[V^b]^\vee
\end{align} \label{th_1}
where $B_n$ are the Bernoulli's numbers. 
\end{lemma}

\begin{lemma} \label{lem_br_eps}
A well known property is $\ad_XX = 0$. A consequence of this is that if the logarithm mapping, $\epsilon = \log(g)$ for a Lie group element $g$, then,
\begin{equation}
    \mathcal{B}_r(\epsilon)\epsilon = \epsilon~ 
\end{equation}
\end{lemma}

\begin{lemma}\cite[sec. 10]{Selig}
The Jacobian $\mathcal{B}_r(X)$ where $X = \begin{bmatrix}\psi^T & q^T\end{bmatrix} = [x]^\vee$, $x \in \mathfrak{se}(3)$ is given as,
\begin{equation} \label{eq_selig_b}
\mathcal{B}_r(X) = \mathbb{I}_{6,6} - \frac{1}{2}\ad_X + \gamma_1(\psi)\ad_X^2 + \gamma_2(\psi)\ad_X^4
\end{equation}
where $\gamma_1(\psi) = \frac{2}{||\psi||^2}+\frac{||\psi|| + 3\sin||\psi||}{4||\psi||(\cos||\psi||-1)}$ and $\gamma_2(\psi) = \frac{1}{||\psi||^4} + \frac{||\psi|| + \sin||\psi||}{4||\psi||^3(\cos||\psi|| - 1)}$. 
\label{lem_exp_jacob}
\end{lemma}

\begin{lemma} \emph{Skew-symmetric property}: For the dynamics of the form \eqref{eq_target_dyn}, the inner product $\langle V_t^b, \ad_{V_t^b}^*\Lambda_t V_t^b \rangle = 0$.\label{lem_passivity}
\end{lemma}

\begin{definition} \label{lem_ad_sim}
An additional operator,  $\ad_h^\sim:\mathfrak{se}(3) \rightarrow \mathfrak{se}(3)^*$ for $h \in \mathfrak{se}(3)^*$ is introduced. For a rigid body with inertia $\Lambda:\mathfrak{se}(3) \rightarrow \mathfrak{se}(3)^*$ and momentum, $h = \begin{bmatrix}h_\omega^T & h_v^T \end{bmatrix}^T = \Lambda V^b \in \mathfrak{se}(3)^*$, $V^b \in \mathfrak{se}(3)$, $\ad_{h}^\sim = \begin{bmatrix}h_{\omega_\times} & h_{p_\times}\\h_{p_\times} & 0\end{bmatrix}$, where $h_\omega$ and $h_v$ are the angular and linear momenta respectively. This can be derived by using the property, $a_\times b  = -b_\times a$, after expanding $\ad_{V_1^b}^*\Lambda V_2^b$ in terms of its rotational and translational components. 

\end{definition}
\begin{lemma}  \label{lem_vel_err}
For a pose $g \in \text{SE(3)}$, and body velocities, $V_1^b$, $V_2^b$, $V_e^b \in \mathfrak{se}(3)$, such that, $V_e^b = V_1^b - \Ad_gV_2^b$, the difference in the coadjoint terms corresponding to velocities $V_1^b$ and $V_2^b$ for inertia $\Lambda$, is given as,
\begin{equation}
    \ad_{V_1^b}^*\Lambda V_1^b - \ad_{\Ad_gV_2^b}^* \Lambda \Ad_g V_2^b = (\ad_{V_1^b}^*\Lambda + \ad_{\Lambda V_1^b}^\sim - \ad_{V_e^b}\Lambda) V_e^b 
\end{equation}
where $\ad^{\sim}_{(.)}$ from Def.~\ref{lem_ad_sim} has been used.

Proof:
\begin{align*}
    \text{L.H.S } =&  \ad_{V_1^b}^*\Lambda(V_e^b + \Ad_g V_2^b) - \ad_{(V_1^b - V_e^b)}^* \Lambda \Ad_g V_2^b \\
    = & \ad_{V_1^b}^*\Lambda V_e^b + \ad_{V_1^b}^*\Lambda\Ad_g V_2^b - \ad_{V_1^b}^* \Lambda \Ad_g V_2^b  + \\ & \ad_{V_e^b}^*\Lambda \Ad_g V_2^b \\
    = & \ad_{V_1^b}^*\Lambda V_e^b + \ad_{V_e^b}^*\Lambda (V_1^b - V_e^b) \\ 
=&  (\ad_{V_1^b}^*\Lambda + \ad_{\Lambda V_1^b}^\sim - \ad_{V_e^b}^*\Lambda) V_e^b 
\end{align*}
\end{lemma}

\begin{lemma}\label{lem_innerProd}
For a pose $g \in \text{SE(3)}$, and body velocities, $V_1^b$, $V_2^b$, $V_e^b \in \mathfrak{se}(3)$, such that, $V_e^b = V_1^b - \Ad_gV_2^b$, the inner product form,
\begin{equation*}
\langle V_e^b, (\ad_{V_1^b}^*\Lambda V_1^b - \ad_{\Ad_gV_2^b}^*\Ad_gV_2^b) - \Lambda \ad_{V_1^b}V_e^b\rangle = 0
\end{equation*}

Proof:
Applying Lemma \ref{lem_vel_err}, to the bracketed term in the right side of the inner product,
L.H.S = 
\begin{equation}
    \begin{split}
        &\langle V_e^b, (\underbrace{\ad_{V_1^b}^*\Lambda + \ad_{\Lambda V_1^b}^\sim - \ad_{V_e^b}^*\Lambda - \Lambda\ad_{V_1^b}}_{\mathcal{C}(V_1^b,V_e^b))})V_e^b \rangle \\
        =&\langle V_e^b, \underbrace{(\ad_{V_1^b}^*\Lambda + \ad_{\Lambda V_1^b}^\sim -  \Lambda\ad_{V_1^b})}_{\tilde{\mathcal{C}}(V_1^b)}V_e^b \rangle \quad \text{(}\because\text{Lemma \ref{lem_passivity})}
    \end{split}
\end{equation}
Using the definitions of $\ad$, $\ad^*$ and $\ad^{\sim}$, it can be verified that $\tilde{\mathcal{C}}(V_1^b)$ is a skew-symmetric matrix. Hence, $\langle V_e^b, \tilde{\mathcal{C}}(V_1^b)V_e^b  \rangle = 0$.
Also a consequence is,
\begin{equation} \label{eq_sum_skew_sym}
    \begin{split}
        &\langle V_e^{b},(P\mathcal{C}(V_1^b,V_e^b) + \mathcal{C}(V_1^b,V_e^b)^TP)V_e^b\rangle=0
    \end{split}
\end{equation}
if $P = \text{diag}(p), ~p  = [p_{1}\mathbb{I}_{3,1}^T,~p_{2}\mathbb{I}_{3,1}^T]^T,~p_{1},p_{2} > 0$.
\QEDB
\end{lemma}
\begin{lemma}\textit{Boundedness} of $V_e^b$: \label{lem_bounded_terms}
For the system defined by \eqref{eq_obs_err_dyn}, $x(0) \in \Omega$, using \eqref{eq_lyap_nsd}, we get $||\epsilon(t)|| < ||\epsilon(0)||$. Since $W(t) \leq \overline{\sigma}({P})||x(0)||^2 \Rightarrow ||x(t)||^2 \leq \frac{\overline{\sigma}({P})}{\underline{\sigma}({P})}||x(0)||^2 =c_2, ~c_2>0 \Rightarrow ~\exists c_1> 0, ||V_e^b(t)|| \leq c_1$.
\end{lemma}

\begin{lemma} \label{lem_cont_ad}
The matrix operators $\ad_{V^b}, ~\ad_{V^b}^*,~\ad_{\Lambda V^b}^{\sim}$ are continuous if $V_b \in \mathfrak{se}(3)$ is bounded such that $||V_b|| < a_1$.

Proof: Applying theorem for continuity and boundedness \cite[th. 2.7-9]{kreyszig1978introductory} for linear operators, $\ad_{V^b}, ~\ad_{V^b}^*,~\ad_{\Lambda V^b}^{\sim}$ are bounded and hence continuous.
\end{lemma} 
\begin{lemma} Boundedness and continuity of $\mathcal{\dot{W}}$: \label{lem_bound_cont}
\begin{enumerate}
    \item For free-floating motion of the Target, $||V_t^b|| < c_4$. \label{it_1}
    \item From \eqref{eq_target_dyn}, $||\dot{V}_t^b||^2 \leq  \overline{\sigma}(\Lambda_t^{-1}\ad_{V_t^b}^*\Lambda_t)||V_t^b||^2$ which proves that $V_t^b \in \mathcal{C}^1$.\label{it_2}
    \item  $\exists c_5 > 0$, such that $||\mathcal{C}(V_t^b,V_e^b)|| < c_5$ after applying Lemma \ref{lem_bounded_terms} for $V_e^b$ and item \ref{it_1} for $V_t^b$. Applying item $2$, and Lemmas \ref{lem_cont_ad} and \ref{lem_bounded_terms} to all the contained $\ad_{(.)}^{(.)}$ terms in $\mathcal{C}(V_t^b,V_e^b)$, we conclude, $\mathcal{C}(V_t^b,V_e^b) \in \mathcal{C}^1$.
    \item If $\mathcal{C}(V_t^b,V_e^b)$ is bounded and continuous, applying Lemma \ref{lem_bounded_terms} to $A(x)$, $Q_1$ is bounded and hence, continuous \cite[th. 2.7-9]{kreyszig1978introductory}. \label{it_4}
    \item Applying, Lemma \ref{lem_bounded_terms} to $Q_2$,  bounded and continuous property of $Q_1 - Q_2$ follows from that of $Q_2$ and item \ref{it_4}.
\end{enumerate}
By applying the above observations, we conclude that $\dot{\mathcal{W}} \in \mathcal{C}^1$ and $\underline{\gamma}||x||^2 \leq |\dot{\mathcal{W}}| \leq \overline{\gamma}||x||^2$, for $\overline{\gamma},~\underline{\gamma} > 0$.
\end{lemma}

\begin{theorem} \emph{Matrosov's theorem}: \label{th_matros}
Consider the system $\dot{x} = f(x,t)$ with $f(t,0) = 0 ~\forall t>0$. Assume there exist two $\mathcal{C}^1$ functions $W(t,x):[0,\infty)\times \Omega \rightarrow \mathbb{R}_+$, $\mathcal{W}(t,x):[0,\infty)\times \Omega \rightarrow \mathbb{R}$ with an open connected set $\Omega \subset \mathbb{R}^n$ containing the origin, a $\mathcal{C}^0$ function $W^*:\Omega\rightarrow\mathbb{R}_+$, three functions exist $\underline{\alpha},\overline{\alpha}(||x||),c \in \mathcal{K}$ such that for every $(x,t) \in \Omega\times [0,\infty)$, 

\begin{enumerate}
    \item $\underline{\alpha}(||x||) \leq W \leq \overline{\alpha}(||x||)$
    \item $\dot{W}(t,x) \leq W^*(x) \leq 0$
    \item $|\mathcal{W}(t,x)|$ is bounded. (auxilliary function) 
    \item max$(d(x,E),|\dot{\mathcal{W}}(t,x)|) \geq c(x)$, where $E = \{x\in \Omega|{W}(x)^*= 0 \}$
    \item $||f(t,x)||$ is bounded.
\end{enumerate}
where $E \equiv \{x \in \Omega| W^*=0\}$
Then:
\begin{enumerate}
    \item $\forall x_0 \in \{x \in \Omega|{W}(t,x) \leq \underline{\alpha}(r) \} ~\forall r>0$ such that a closed ball $B_r \subset \Omega$, $x(t) \rightarrow 0$ uniformly in $t_0$ as $t\rightarrow\infty$.
    \item The origin is uniformly asymptotically stable.
\end{enumerate}
\end{theorem}
\begin{lemma} \label{lem_matros}
In theorem \ref{th_matros}, condition $4$ is satisfied for the following:
\begin{enumerate}
    \item $\dot{\mathcal{W}}(t,x)$ is continuous in both arguments and depends on time in the following way, $\dot{\mathcal{W}}(t,x) = g(x,\beta(t))$ where $g$ is continuous in both its arguments. $\beta$ is also continuous and its image lies in a bounded set $K_1$.
    \item $\exists$ class $\mathcal{K}$ function, $k$, such that $|\dot{\mathcal{W}}(t,x)| \geq k(||x||) ~\forall x\in E, ~t>0$ .
\end{enumerate}
\end{lemma}

\bibliographystyle{unsrt}
\bibliography{bibFile}{}

%

\end{document}